%% file: main.tex
\documentclass[10pt,twocolumn,letterpaper]{article}

\usepackage{cvpr}              % To produce the CAMERA-READY version
\input{preamble}

\definecolor{cvprblue}{rgb}{0.21,0.49,0.74}
\usepackage[pagebackref,breaklinks,colorlinks,citecolor=cvprblue]{hyperref}

\def\paperID{*****} % *** Enter the Paper ID here
\def\confName{CVPR}
\def\confYear{2024}

\title{Agriculture-Vision Challenge 2024 – The Runner-Up Solution for Agricultural
Pattern Recognition via Class Balancing and Model Ensemble}

\author{Wang Liu\\
Hunan University\\
Changsha, Hunan, China\\
{\tt\small liuwa@hnu.edu.cn}
\and
Zhiyu Wang\\
Hunan University\\
Changsha, Hunan, China\\
{\tt\small wangzhiyu.wzy1@gmail.com}
\and
Puhong Duan\\
Hunan University\\
Changsha, Hunan, China\\
{\tt\small puhong\_duan@hnu.edu.cn}
\and
Zhuojun Xie\\
Hunan University\\
Changsha, Hunan, China\\
{\tt\small xzj22@hnu.edu.cn}
\and
Xudong Kang\\
Hunan University\\
Changsha, Hunan, China\\
{\tt\small xudong\_kang@163.com}
\and
Shutao Li\\
Hunan University\\
Changsha, Hunan, China\\
{\tt\small shutao\_li@hnu.edu.cn}
}

\begin{document}
\maketitle
\input{sec/0_abstract}
\input{sec/1_intro-my}

\bibliographystyle{ieeenat_fullname}
\bibliography{main}


\end{document}

%% file: preamble.tex
\usepackage[dvipsnames]{xcolor}

%% file: sec/0_abstract.tex
\begin{abstract}
The Agriculture-Vision Challenge at CVPR 2024 aims at leveraging semantic segmentation models
to produce pixel-level semantic segmentation labels within regions of interest for multi-modality satellite images.
It is one of the most famous and competitive challenges for global researchers
to break the boundary between computer vision and agriculture sectors.
However, there is a serious class imbalance problem in the agriculture-vision dataset,
which hinders the semantic segmentation performance. To solve this problem,
firstly, we propose a mosaic data augmentation with a rare class sampling strategy to enrich long-tail class samples.
Secondly, we employ an adaptive class weight scheme to suppress the contribution
of the common classes while increasing the ones of rare classes.
Thirdly, we propose a probability post-process to increase the predicted value of the rare classes.
Our methodology achieved a mean Intersection over Union (mIoU) score
of 0.547 on the test set, securing second place in this challenge.
\end{abstract} 

%% file: sec/1_intro-my.tex
\section{Introduction}
\label{sec:intro}
In the field of remote sensing agriculture, semantic segmentation plays a pivotal role in automatically classifying and delineating areas of interest, such as crop types, diseased vegetation, and land cover, from satellite and aerial imagery \cite{chiu2020agriculture, Chiu_2020_CVPR}. In recent years, semantic segmentation research has been in full swing. However, the effectiveness and efficiency of these methods are still not satisfactory for use in agriculture recognition. Thus, the 4th Agriculture-Vision Prize Challenge aims to encourage research in developing novel and effective algorithms for agricultural pattern recognition from aerial images.

\begin{figure}[t]
    \centering
    \vspace{0.2cm}
    \includegraphics[width=3.2in, keepaspectratio]{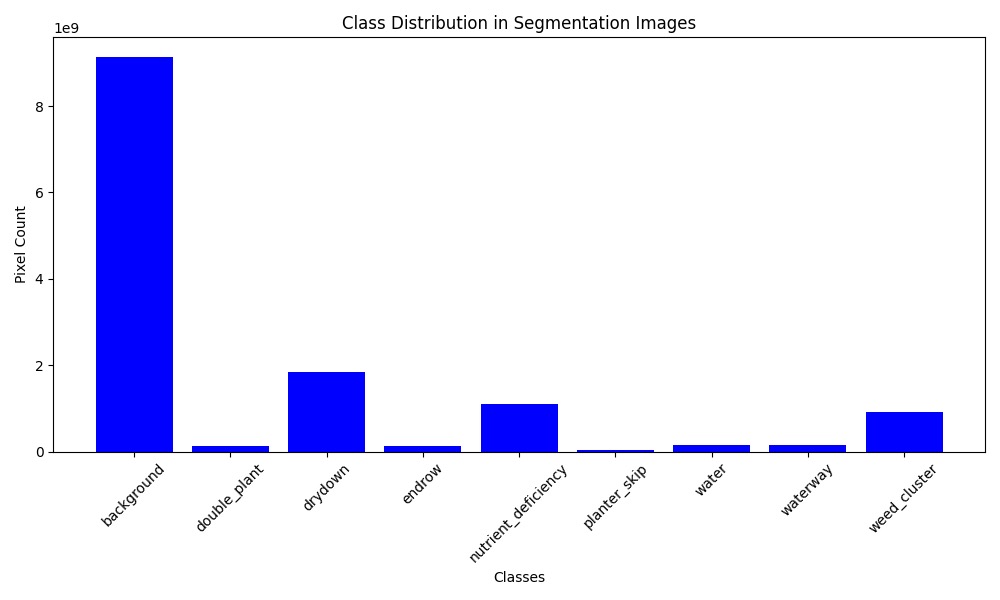}
    \vspace{-0.4cm}
    \caption{The number of pixels in each class.}
    \label{fig-class-dist}
\end{figure}
A major challenge persists due to the severe class imbalance present in the Agriculture-vision dataset used for training segmentation models. Typically, some classes are overly represented while others are scarce, leading to models that are biased towards more frequent categories and perform poorly on rarer ones. This imbalance hampers the overall accuracy and applicability of segmentation technologies in practical agricultural settings, where diverse and balanced detection capabilities are crucial. To address this issue, this paper explores novel approaches designed specifically to mitigate class imbalance in dataset preparation and model training. %We introduce enhanced algorithms that aim to equalize the influence of all classes, thereby improving the robustness and generalization of the models to various remote-sensing agricultural scenarios.% This work not only contributes to the development of more precise segmentation methods but also broadens the scope of their application, ultimately aiding in more effective monitoring and management of agricultural resources.

\section{Method}

\subsection{Data Analysis}

The Agriculture-Vision dataset consists of RGB-NIR aerial images of farmlands across the US captured between 2017 and 2019 with a resolution of up to 10 cm/pixel. After the official preprocessing, it contains 94,986 aerial farmland images. Each image has 512$\times$512 pixels. A total of 8 patterns commonly found in farmlands- including weeds, end-rows, nutrient deficiency, and others are carefully annotated by agronomy experts. The number of pixels in each class is shown as Fig. \ref{fig-class-dist}. It can be seen that the background category accounts for a too high proportion. There is a significant class imbalance within this dataset. This class imbalance problem presents a challenge as it may skew the performance of segmentation models, making them biased toward the majority class while neglecting the minority class.

\subsection{Class Balancing}
To solve this problem, we address it from the three aspects including data pre-process, loss function, and probability post-process.

For the class-balance data pre-process, we calculate the sampling probability of each class based on the frequency of all classes in the dataset so that the long-tail classes are sampled more often \cite{DAFormer}. However, directly adding the sampled images to the training set is prone to overfitting problems in long-tail categories. To alleviate this problem, we introduce mosaic augmentation to diversify the rare class samples.

For the class-balance loss function, we calculate the weight of each category based on the number of samples of each class that have participated in the training \cite{ACWLoss}. The fewer the number of categories, the greater the weight. This allows rare classes to participate more in model training to alleviate the problem of sample imbalance.

We propose a class-balance post-process. After the model outputs a probability value for each class, we directly multiply the probability of the rare class by 2.0 and the probability of the background class by 0.95. This method can improve the predicted probability of the rare class and prevent it from being overwhelmed by the background class.

\subsection{Final Solution}

%\begin{figure}[t]
%    \centering
%    \includegraphics[width=3.2in, keepaspectratio]{./Figures/RCS_mosaic.png}
%    \vspace{-0.2cm}
%    \caption{This data flow of our final solution.}
%    \label{fig-solution}
%    \vspace{-0.2cm}
%\end{figure}

\begin{figure}[t]
    \centering
    \includegraphics[width=3.2in, keepaspectratio]{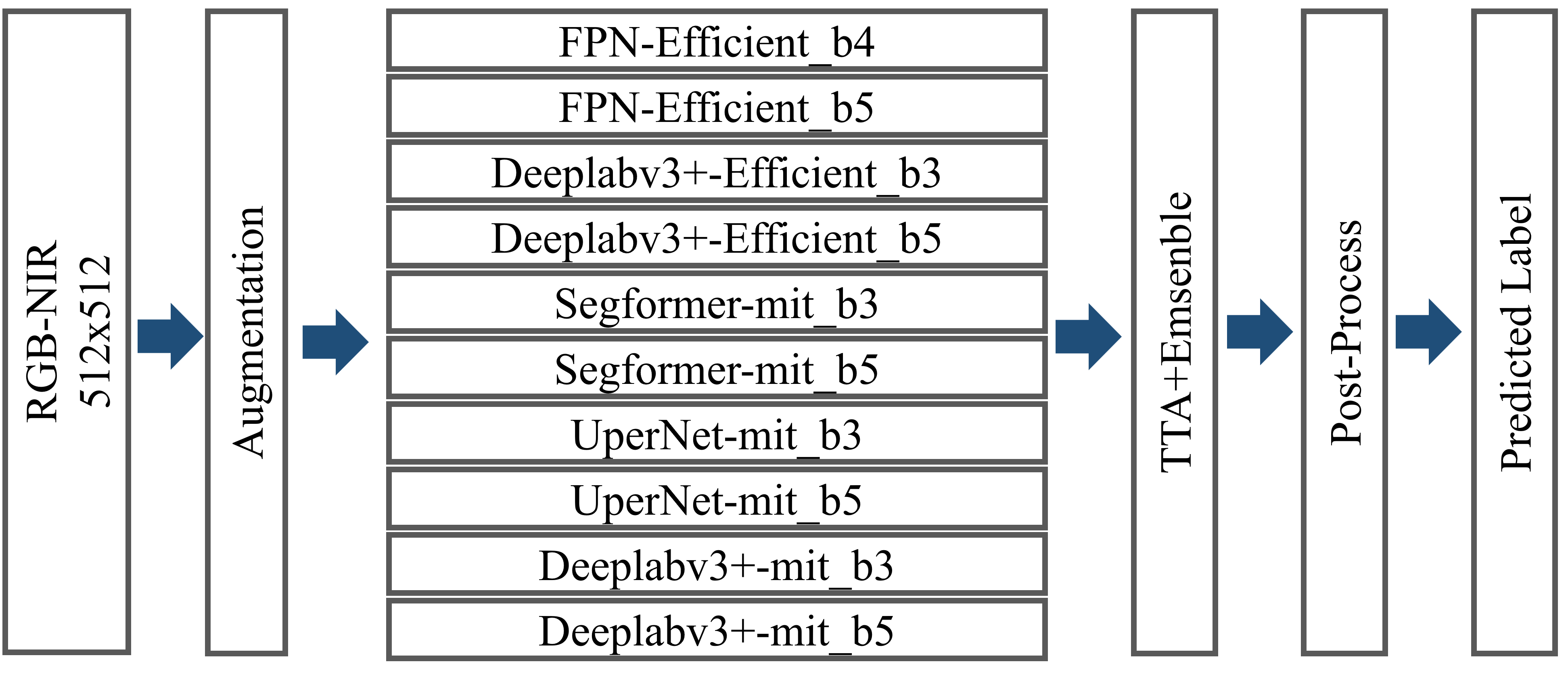}
    \vspace{-0.2cm}
    \caption{This data flow of our final solution.}
    \label{fig-solution}
    \vspace{-0.2cm}
\end{figure}

Many images have multiple labels, and even have overlapped labels. To simplify the training process, we replace the overlapped labels with single ones. The final solution of this work is illustrated in Fig. \ref{fig-solution}. We concatenate the RGB and NIR images as a four-channel input. The data augmentations are employed to enrich the training set. We choose FPN \cite{FPN}, DeeplabV3+ \cite{DeeplabV3plus}, Segformer \cite{Segformer}, and UperNet \cite{UperNet} as the basic model architecture with MixVisionTransformer (Mit) \cite{Segformer} and EfficientNet \cite{EfficientNet}. The test time augmentation (TTA) is utilized to improve the performance of each single model. After that, we utilize the math average as the basic ensemble to fuse the class probabilities predicted by these single models. Finally, the post-process is introduced to refine the ensemble results.

\section{Experiments}

\subsection{Implement Details}
All the experiments are conducted with a single NVIDIA RTX 4090-24GB GPU on Ubuntu 22.04 LTS OS. The proposed method is implemented based on PyTorch-1.8.1+cu111. In this study, the "Adamw" is used as an optimizer with a weight decay of 0.01. The batch size is set to 16. The learning rate is initialized to 6e-6, and its descent strategy is a "poly" policy formulated as ${\left( {1 - \frac{{iter}}{{\max \_iter}}} \right)^{power}}$, $power=0.9$. We use the warm-up strategy before the learning rate descends. Additionally, the model is trained with 160000 iterations. Several widely used data augmentation strategies, i.e., random resize and crops, random horizontal and vertical flips, random rotation, and colorjitter are utilized. The cropping size is set as $512\times512$. The mIoU is utilized as the evaluating metric.

\subsection{Result and Analysis}

\begin{table}[t!]
\caption{Analysis of the effectiveness of the components with Seformer-Mit-b3 as the baseline method.} \vspace{-0.3cm}
\label{table-abl-propose}
\centering\normalsize\renewcommand\arraystretch{0.9}
\begin{tabular}{c|c}
\hline
\hline
Methods                         & mIoU \\
\hline
baseline                        & 0.452 \\
baseline+ACWLoss                & 0.492 \\
baseline+ACWLoss+RCS\_Mosaic    & 0.516 \\
baseline+ACWLoss+RCS\_Mosaic+Post  & 0.524 \\
\hline
\hline
\end{tabular}
\vspace{-0.4cm}
\end{table}

The results to analyze the effectiveness of the component are listed in Table \ref{table-abl-propose}. It can be seen that the recognition accuracy is improved a lot after utilizing the data pre-process, loss function, and probability post-process. To analyze the effect of the model architectures and backbones, we select UperNet \cite{UperNet}, DeeplabV3+ \cite{DeeplabV3plus}, FPN \cite{FPN}, and Segformer \cite{Segformer} as the comparing architecture, and ResNet \cite{ResNet}, LSKNet \cite{LSKNet}, Mit \cite{Segformer}, and EfficientNet \cite{EfficientNet} as the comparing backbones. Table \ref{table-abl-model} presents the results. It can be seen that the Mit and EfficientNet backbones can yield better performance than others. The model architecture does not have much impact on the results. We ensemble the models that are signed by '*' and use the post-process to refine the ensemble probability resulting in the best score.

\begin{table*}[t!]
\caption{Analysis of the effectiveness of semantic segmentation architectures and backbones. EffNet indicates EfficientNet.}
\small
\label{table-abl-model}
\centering\normalsize\renewcommand\arraystretch{0.9}
\begin{tabular*}{\hsize}{@{\extracolsep{\fill}}c|c|ccccccccc|c}
\hline
\hline
Architectures & Backbones    & BG    & DP    & DR    & EN    & ND    & PS    & WA    & WW    & WC    & mIoU \\
\hline
UperNet       & ResNet-50d8  & 0.717 & 0.385 & 0.553 & 0.312 & 0.443 & 0.352 & 0.717 & 0.420 & 0.338 & 0.471 \\
UperNet       & LSKNet-tiny  & 0.712 & 0.382 & 0.565 & 0.363 & 0.459 & 0.396 & 0.727 & 0.399 & 0.346 & 0.483 \\
UperNet       & LSKNet-small & 0.714 & 0.394 & 0.568 & 0.361 & 0.461 & 0.402 & 0.733 & 0.402 & 0.356 & 0.488 \\
UperNet*      & Mit-b3       & 0.741 & 0.427 & 0.588 & 0.377 & 0.503 & 0.404 & 0.740 & 0.445 & 0.390 & 0.513 \\
UperNet*      & Mit-b5       & 0.744 & 0.430 & 0.590 & 0.387 & 0.509 & 0.429 & 0.741 & 0.440 & 0.395 & 0.518 \\
DeeplabV3+    & ResNet-50d8  & 0.719 & 0.386 & 0.555 & 0.321 & 0.446 & 0.367 & 0.725 & 0.392 & 0.339 & 0.472 \\
DeeplabV3+    & Swin-base    & 0.737 & 0.426 & 0.581 & 0.371 & 0.484 & 0.370 & 0.748 & 0.407 & 0.376 & 0.500 \\
DeeplabV3+    & Swin-large   & 0.743 & 0.437 & 0.584 & 0.398 & 0.500 & 0.386 & 0.763 & 0.417 & 0.386 & 0.513 \\
DeeplabV3+*   & Mit-b3       & 0.739 & 0.428 & 0.583 & 0.389 & 0.506 & 0.433 & 0.746 & 0.446 & 0.378 & 0.516 \\
DeeplabV3+*   & Mit-b5       & 0.743 & 0.437 & 0.592 & 0.393 & 0.506 & 0.399 & 0.741 & 0.428 & 0.390 & 0.515 \\
DeeplabV3+*   & EffNet-b3    & 0.749 & 0.430 & 0.585 & 0.392 & 0.498 & 0.401 & 0.751 & 0.438 & 0.375 & 0.513 \\
DeeplabV3+*   & EffNet-b5    & 0.752 & 0.442 & 0.585 & 0.409 & 0.486 & 0.396 & 0.747 & 0.429 & 0.381 & 0.514 \\
FPN*          & EffNet-b4    & 0.751 & 0.439 & 0.576 & 0.409 & 0.494 & 0.410 & 0.748 & 0.428 & 0.368 & 0.514 \\
FPN*          & EffNet-b5    & 0.753 & 0.447 & 0.577 & 0.417 & 0.497 & 0.427 & 0.749 & 0.416 & 0.386 & 0.519 \\
Segformer*    & Mit-b3       & 0.740 & 0.432 & 0.589 & 0.381 & 0.503 & 0.408 & 0.756 & 0.439 & 0.387 & 0.516 \\
Segformer*    & Mit-b5       & 0.745 & 0.431 & 0.583 & 0.399 & 0.512 & 0.418 & 0.748 & 0.460 & 0.373 & 0.519 \\
\hline
Ensemble+Post & Mit-b5       & {\bf 0.753} & {\bf 0.469} & {\bf 0.604} & {\bf 0.440} & \bf{0.525} & {\bf 0.461} & {\bf 0.769} & \bf{0.478} & {\bf 0.417} & \bf{0.547}\\
\hline
\hline
\end{tabular*}
\end{table*}

\section{Conclusion}
In this paper, we propose our solution to the 4th Agriculture-Vision Challenge in CVPR 2024.
We propose a mosaic augmentation with rare class sampling making the number of long-tail category samples enriched while the diversity of the dataset has been expanded.
We introduce an adaptive loss reweighting scheme. It suppresses the excessive participation of common classes in training, while the contribution of long-tail classes to network optimization has also been improved.
We design a probability post-process strategy. The mean accuracy in all classes has been improved.
We finally accomplished a mIoU of 0.547, achieving 2nd place in this challenge.